# Comparisons of Reasoning Mechanisms for Computer Vision

Ze-Nian Li

*Department of Electrical Engineering and Computer Science*
*University of Wisconsin-Milwaukee*

ABSTRACT

An evidential reasoning mechanism based on the Dempster-Shafer theory of evidence is introduced. Its performance in real-world image analysis is compared with other mechanisms based on the Bayesian formalism and a simple weight combination method.

## 1. Introduction

A pyramid vision system that handles real-world image analysis in a parallel and hierarchical manner has been developed [3, 5]. Its evidential reasoning mechanism is an adaptation and extension of the Dempster-Shafer (D-S) theory of evidence. Some object recognition examples using this evidential reasoning mechanism have been illustrated in our previous paper [4]. An immediate question is: "Does it work better than other reasoning mechanisms?" We are conducting experimental comparisons of various reasoning mechanisms for real-world image analysis. In this paper empirical comparisons are made among reasoning mechanisms using the D-S theory, the Bayesian formalism, and a simple weight combination method.

## 2. Reasoning Based on the Dempster-Shafer Theory of Evidence

### 2.1. The Reasoning Mechanism

In [3, 4] a modular evidential reasoning mechanism based on the D-S set-theoretical theory of evidence [1, 6] is presented. It was shown that the new reasoning mechanism can be used effectively in a massively parallel hierarchical pyramid vision system, characterized by the use of key features. The evidential reasoning mechanism has the following characteristics:

(1) **Evidence Accumulation.** Each extracted feature is viewed as a piece of evidence. Belief functions of independently extracted features are combined using Dempster's combination rule. The accumulated evidence is represented by a belief function whose mass distribution function is $m_e$.

(2) **A Modular Knowledge Representation Technique.** The system's knowledge of a hypothetical object is represented by a belief function whose mass distribution function is $m_s$. Probability mass values are associated with the expected feature components of the hypothesized objects. The importance of the feature is represented by the amount of the mass assigned to it. The reasoning process is able to utilize small and modular pieces of knowledge and to take advantage of the use of *key features*.

(3) **Hypothesis Verification.** A new belief function $Bel(o)$ is introduced to represent the result of the hypothesis verification.

The Mass Function of *Bel* (*o*)

(a) $m(o) = \sum_{\substack{A_i \subseteq B_j \\ B_j \neq \Theta}} m_e(A_i) m_s(B_j),$

(b) $m(\Theta) = 1 - m(o).$

The belief function thus generated is a simple belief function that represents the belief for the hypothesized object *o*. The use of the set inclusion operator in the definition of *Bel* (*o*) performs the consistency check between the system's knowledge and the accumulated evidence. The creation of this new belief function has extended the notations of the belief functions and the Dempster combination rule. In our pyramid vision system the evidential reasoning is applied hierarchically. The *Bel* (*o*) can be used, for example, together with other pieces of evidence by a higher level hypothesis verification process in the pyramid. See [3, 4] for a more complete description of this reasoning mechanism.

## 2.2. Test Result Based on the D-S Method

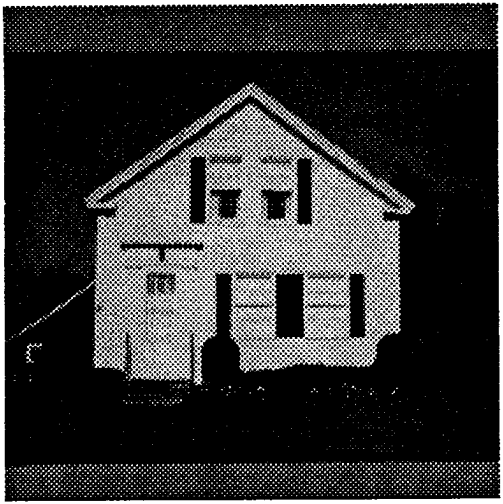
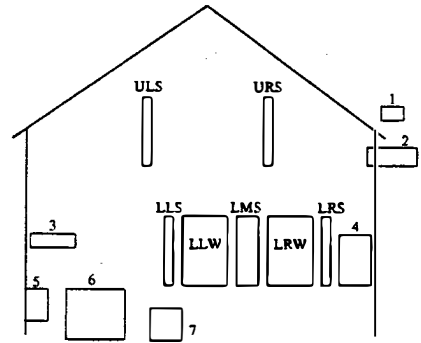

ULS — Upper Left Shutter  URS — Upper Right Shutter
LLS — Lower Left Shutter  LMS — Lower Middle Shutter
LRS — Lower Right Shutter
LLW — Lower Left Window  LRW — Lower Right Window

Fig. 1  Test Image 'House.sri'    Fig. 2  Possible Window or Shutter Areas

This section illustrates a D-S evidential reasoning example with the image 'House.sri' (Fig. 1), i.e., how the system recognizes the 'window-assemblies' (windows and shutters).

Using the pairs of vertical long edges as key features for windows and shutters, 14 possible window or shutter areas are located [3]. Figure 2 is a graphical display of these 14 areas. For ease of illustration real window and shutter areas in 'House.sri' are labeled with 'LLW', 'LLS', etc. Others are labeled from 1 to 7. (The program is not able to find the top windows at this time.)

Similar to what we did in [4], features of *elongation*, *texture* and *boundary* of the areas are examined. A feature is compared to the **typical** feature value of a hypothesized object. A *Bel* value is assigned according to the feature's 'goodness'. The simple belief functions ( *Bel* (*elong* (*sht* )), *Bel* (*text* ), etc. ) are then combined to derive a new belief function. For *shutter* it is

$$Bel(elong(sht)) \oplus Bel(elong(non-sht)) \oplus Bel(text) \oplus Bel(top-bound) \oplus Bel(bottom-bound),$$

whose mass distribution is denoted by $m_e(sht)$. In the same way, $m_e(wnd)$ is derived.

The mass distributions for the knowledge sources used to verify the shutter or window hypotheses are:

$m_{s_1}(sht)$:  $m(elong(sht)) = 0.4$,  $m(text) = 0.15$,
          $m(boundary) = 0.25$,  $m(\Theta) = 0.2$;

$m_{s_1}(wnd)$:  $m(elong(wnd)) = 0.4$,  $m(text) = 0.15$,
          $m(boundary) = 0.25$,  $m(\Theta) = 0.2$.

Since the 'elongation' of the area is an important feature used to discriminate shutters from windows in this example, it is emphasized by assigning a large portion of the total mass to the proposition *elong*. The resulting belief functions are $Bel(sht)$ and $Bel(wnd)$ as shown in Table 1.

Table 1  Belief Values for Features in 'House.sri'

|  | Possible window or shutter areas | | | | | | | | | | | | | |
| --- | --- | --- | --- | --- | --- | --- | --- | --- | --- | --- | --- | --- | --- | --- |
|  | ULS | URS | LLS | LMS | LRS | LLW | LRW | 1 | 2 | 3 | 4 | 5 | 6 | 7 |
| $Bel(elong(sht))$ | 0.5 | 0.5 | 0.5 | 0.3 | 0.5 | 0 | 0 | 0 | 0.3 | 0.3 | 0 | 0 | 0 | 0 |
| $Bel(elong(wnd))$ | 0 | 0 | 0 | 0.3 | 0 | 0.5 | 0.5 | 0.5 | 0.3 | 0.3 | 0.5 | 0.5 | 0.5 | 0.5 |
| $Bel(text)$ | 0.4 | 0.4 | 0.4 | 0.4 | 0.4 | 0.4 | 0.4 | 0 | 0 | 0 | 0.4 | 0.4 | 0.4 | 0.4 |
| $Bel(top-bound)$ | 0.3 | 0.6 | 0.6 | 0.6 | 0.6 | 0.6 | 0.6 | 0.1 | 0.3 | 0 | 0 | 0 | 0.3 | 0.1 |
| $Bel(btm-bound)$ | 0.3 | 0.6 | 0 | 0.6 | 0.6 | 0.3 | 0.1 | 0.1 | 0.1 | 0 | 0 | 0 | 0.1 | 0.3 |
| $Bel(sht)$ | .388 | .470 | .410 | .357 | .470 | .240 | .220 | .048 | .183 | .092 | .060 | .060 | .154 | .154 |
| $Bel(wnd)$ | .188 | .270 | .210 | .357 | .270 | .440 | .420 | .248 | .183 | .092 | .260 | .260 | .354 | .354 |
| $Bel(h-sibl)$ | 0.6 | 0.6 | 0.6 | 0.6 | 0.6 | 0.6 | 0.6 | 0 | 0 | 0 | 0.6 | 0 | 0 | 0 |
| $Bel'(sht)$ | .451 | .509 | .467 | .429 | .509 | .348 | .334 | .034 | .128 | .064 | .222 | .042 | .108 | .108 |
| $Bel'(wnd)$ | .311 | .369 | .327 | .429 | .369 | .488 | .474 | .174 | .128 | .064 | .362 | .182 | .248 | .248 |

In house scenes windows and shutters in 'window-assemblies' are usually horizontal siblings. The belief function $Bel(h-sibl)$ is introduced to represent this geometrical relation. By incorporating this into the reasoning process, the final belief functions are $Bel'(sht)$ and $Bel'(wnd)$.

### 2.3. Methods for Decision Making

For this example the system is expected to identify the most probable window and shutter areas. At least two simple methods may be recommended for decision making.

Method 1: Choose the label with the highest *Bel* value.

For each possible window or shutter area, examine its $Bel'(sht)$, $Bel'(wnd)$, and $Bel'(others)$. Although the $Bel'(others)$ are not shown in the table, a rough assumption will be made for this discussion that $Bel'(others)$ is low for a true window or shutter area, and high for other areas. It can be seen from Table 1 that all the areas would be labeled correctly except the LMS will end up in a draw between *sht* and *wnd*.

Method 2: Choose a threshold ($\theta$) for each label.

An area will be labeled as a window or shutter if its associated belief value is higher than the chosen threshold. As shown in Table 1, the real shutter areas obtain the highest $Bel'(sht)$ values (0.451, 0.509, 0.467, 0.429, and 0.529); the real window areas have the highest $Bel'(wnd)$ values (0.488, 0.474). Therefore, it is not difficult to assign a unique label to each area. In case the initial threshold was too low, e.g., 0.4 for both *sht* and *wnd*, the area LMS would be labeled both as *sht* and *wnd*. An adjustment would be needed to raise some of the thresholds. For instance, if $\theta(wnd)$ is raised to about 0.45, then the unique decision would be made.

## 3. Reasoning Based on the Bayesian Formalism

For empirical comparison an experiment was run by replacing our D-S evidential reasoning mechanism with a method based on the Bayesian formalism.

### 3.1. The Reasoning Mechanism

The same set of features (*elongation*, *texture*, etc.) is employed. The reasoning process will compute the posterior probability for the hypothesized object ($H$) given multiple uncertain evidence ($E_i$'s). The mechanism for updating Bayesian probabilities suggested by Duda, Hart and Nilsson [2] is adopted.

The *prior odds* on a hypothesis $H$ is defined to be

$$O(H) = \frac{P(H)}{P(\bar{H})} = \frac{P(H)}{1 - P(H)}, \quad (1)$$

and the *posterior odds* to be

$$O(H|E) = \frac{P(H|E)}{P(\bar{H}|E)} = \lambda O(H), \quad (2)$$

when the evidence $E$ is known to be true, and the *likelihood ratio* $\lambda$ is defined as

$$\lambda = \frac{P(E|H)}{P(E|\bar{H})}. \quad (3)$$

In a strictly analogous fashion, if $E$ is known to be false, then

$$O(H|\bar{E}) = \frac{P(H|\bar{E})}{P(\bar{H}|\bar{E})} = \bar{\lambda} O(H), \quad (4)$$

where $\bar{\lambda}$ is defined as

$$\bar{\lambda} = \frac{P(\bar{E}|H)}{P(\bar{E}|\bar{H})}. \quad (5)$$

If an uncertain evidence $E_i'$ is observed, then $P(E_i|E_i')$ is used to represent the probability that $E_i$ is true under the observation of $E_i'$. It was assumed in [2] that

$$P(H|E_i') = P(H|E_i) P(E_i|E_i') + P(H|\bar{E}_i) P(\bar{E}_i|E_i'). \quad (6)$$

For combining multiple uncertain evidence an effective likelihood ratio $\lambda_i'$ is defined for each single feature by

$$\lambda_i' = \frac{O(H|E_i')}{O(H)}. \quad (7)$$

Assuming $E_i$'s are conditionally independent, the *posterior odds* given $E_1', \cdots, E_n'$ is

$$O(H|E_1', \cdots, E_n') = \left[\prod_{i=1}^{n} \lambda_i'\right] O(H). \quad (8)$$

The reasoning process works as follows. The *prior odds* $O(H)$ for possible objects (*window*, *shutter*, or *others* in this example) should be given. For each observed (uncertain) feature $E_i'$ (e.g., *elongation*, *texture*, etc.), the $\lambda_i$ and $\bar{\lambda}_i$ are also given. The system will need some subjective knowledge for these $\lambda_i$ and $\bar{\lambda}_i$. Alternatively, they may be generated from previous statistical data. The $O(H|E_i)$ and $O(H|\bar{E}_i)$ can be derived from Eqs. (2) and (4). The $P(H|E_i)$ and $P(H|\bar{E}_i)$ can consequently be obtained. ($P = \frac{O}{O+1}$ and $O = \frac{P}{1-P}$, the conversions between $P$ and $O$ are used here and thereafter.) For each observed feature the initial belief value $Bel_i$ in D-S method is now simply used as $P(E|E_i')$, which will enable simple comparison between these two methods later on. It follows that $P(\bar{E}|E_i') = 1 - P(E|E_i')$. With Eq. (6) the $P(H|E_i')$ is calculated and will then be used to derive $O(H|E_i')$. Subsequently, $\lambda_i'$ for feature $i$ is calculated from (7). After obtaining all $\lambda_i$'s the *posterior odds* $O(H|E_1', \cdots, E_n')$ can be computed from (8). Finally, the $P(H|E_1', \cdots, E_n')$ for each possible object can be derived.

**Example:** The impact of the observed *texture* to the hypothesized object *window*.

Let us consider a small set of possible objects $\{wnd, sht, others\}$. Suppose the prior probabilities $P(wnd) = 0.167$, $P(sht) = 0.333$, and $P(others) = 0.5$. For the feature *texture*, $\lambda_{text} = 2$ and $\bar{\lambda}_{text} = 0.8$. Assume that the observation on texture has 40% certainty that there is some window texture, i.e., $P(text|text') = 0.4$.

$$O(wnd) = \frac{0.167}{1 - 0.167} = 0.2.$$

$$O(wnd|text) = 2 \times 0.2 = 0.4, \quad P(wnd|text) = \frac{0.4}{1 + 0.4} = 0.286,$$

$$O(wnd|\overline{text}) = 0.8 \times 0.2 = 0.16, \quad P(wnd|\overline{text}) = \frac{0.16}{1 + 0.16} = 0.138,$$

$$P(wnd|text') = 0.286 \times 0.4 + 0.138 \times (1 - 0.4) = 0.197,$$

$$O(wnd|text') = \frac{0.197}{1 - 0.197} = 0.245,$$

$$\lambda_{text}' = \frac{0.245}{0.2} = 1.227.$$

Apparently, similar computations can be made to obtain $\lambda_i$'s for other observed features (*elong*, *top-bound*, *btm-bound*, and *h-sibl*). Hence, the final updated probabilities can be derived from Eq. (8). Table 2 shows the result for reasoning with this implementation of Bayesian formalism. The first six rows are $P(E|E_i')$s for observed features. The posterior probabilities after the combination of the multiple evidence are denoted by $P'(wnd)$ and $P'(sht)$ respectively.

Table 2  Probability Values for Features in 'House.sri'

| | Possible window or shutter areas | | | | | | | | | | | | | |
|---|---|---|---|---|---|---|---|---|---|---|---|---|---|---|
| | ULS | URS | LLS | LMS | LRS | LLW | LRW | 1 | 2 | 3 | 4 | 5 | 6 | 7 |
| $P(elong(sht))$ | 0.5 | 0.5 | 0.5 | 0.3 | 0.5 | 0 | 0 | 0 | 0.3 | 0.3 | 0 | 0 | 0 | 0 |
| $P(elong(wnd))$ | 0 | 0 | 0 | 0.3 | 0 | 0.5 | 0.5 | 0.5 | 0.3 | 0.3 | 0.5 | 0.5 | 0.5 | 0.5 |
| $P(text)$ | 0.4 | 0.4 | 0.4 | 0.4 | 0.4 | 0.4 | 0.4 | 0 | 0 | 0 | 0.4 | 0.4 | 0.4 | 0.4 |
| $P(top-bound)$ | 0.3 | 0.6 | 0.6 | 0.6 | 0.6 | 0.6 | 0.6 | 0.1 | 0.3 | 0 | 0 | 0 | 0.3 | 0.1 |
| $P(btm-bound)$ | 0.3 | 0.6 | 0 | 0.6 | 0.6 | 0.3 | 0.1 | 0.1 | 0.1 | 0 | 0 | 0 | 0.1 | 0.3 |
| $P(h-sibl)$ | 0.6 | 0.6 | 0.6 | 0.6 | 0.6 | 0.6 | 0.6 | 0 | 0 | 0 | 0.6 | 0 | 0 | 0 |
| $P'(sht)$ | .552 | .708 | .541 | .641 | .708 | .430 | .371 | .084 | .166 | .107 | .201 | .094 | .148 | .148 |
| $P'(wnd)$ | .211 | .345 | .195 | .495 | .345 | .487 | .419 | .090 | .087 | .050 | .221 | .099 | .166 | .166 |

For the above computation the prior probabilities are chosen to be $P(wnd) = 0.167$, $P(sht) = 0.333$, and $P(others) = 0.5$. The reasons for this are that: (1) For the house pictures that we are analyzing, each window has two shutters. Thus, prior $P(sht) = 2 \times P(wnd)$. (2) Usually, after initial image analysis steps the number of hypothesized possible window or shutter areas is much larger than the the number of actual window or shutter areas. Therefore, $P(others)$ is assigned a large prior probability.

## 3.2. Analysis on the Result

The result is comparable with the D-S method. All five shutter areas have higher $P'(sht)$ than their $P'(wnd)$ (e.g., 0.552 > 0.221 for ULS), and two window areas have higher $P'(wnd)$ than their $P'(sht)$ (e.g., 0.487 > 0.430 for LLW). The correct decision would be made if decision method 1 is the choice. However, there would be some problem if decision method 2 is used. Notice that area LMS has an unexpectedly high $P'(wnd)$ (0.495) which is even higher than the $P'(wnd)$ values of the two true window areas. It is not apparent how the system can avoid identifying LMS as a *wnd* when using decision method 2.

As expected, the values of the $\lambda_i$ and $\bar{\lambda}_i$ will have some impact on the posterior probabilities. We experimented with many different groups of these values. Although the magnitudes of the posterior probabilities change substantially, the relative measure of these probabilities is not significantly affected. Namely, the areas having higher $P$'s always have higher $P$'s, and vice versa. Noticeably, the $P'(wnd)$ for LMS is always higher than the $P'(wnd)$ for LRW.

## 4. A Simple Weight Combination Method

An experiment was also run by using a simple weight combination method which is of practical use in some of the perception systems. The initial belief values used in the D-S method are now simply treated as *weights* (WT). Weights for the features that support the proposition, e.g., shutter, are summed up. Weights for the features that support the negation of the proposition are subtracted from the sum. Thus for shutters,

$$WT(sht) = WT(elong(sht)) - WT(elong(\overline{sht})) + WT(text)$$
$$+ WT(top-bound) + WT(bottom-bound).$$

With the consideration of the possible support from siblings,

$$WT'(sht) = WT(sht) + WT(h-sibl).$$

In the same way, $WT(wnd)$ and $WT'(wnd)$ are obtained.

Table 3 lists all these weights for 'House.sri'.

Table 3 Weights for Features in 'House.sri'

| | Possible window or shutter areas | | | | | | | | | | | | | |
|---|---|---|---|---|---|---|---|---|---|---|---|---|---|---|
| | ULS | URS | LLS | LMS | LRS | LLW | LRW | 1 | 2 | 3 | 4 | 5 | 6 | 7 |
| $WT(elong(sht))$ | 0.5 | 0.5 | 0.5 | 0.3 | 0.5 | 0 | 0 | 0 | 0.3 | 0.3 | 0 | 0 | 0 | 0 |
| $WT(elong(wnd))$ | 0 | 0 | 0 | 0.3 | 0 | 0.5 | 0.5 | 0.5 | 0.3 | 0.3 | 0.5 | 0.5 | 0.5 | 0.5 |
| $WT(text)$ | 0.4 | 0.4 | 0.4 | 0.4 | 0.4 | 0.4 | 0.4 | 0 | 0 | 0 | 0.4 | 0.4 | 0.4 | 0.4 |
| $WT(top-bound)$ | 0.3 | 0.6 | 0.6 | 0.6 | 0.6 | 0.6 | 0.6 | 0.1 | 0.3 | 0 | 0 | 0 | 0.3 | 0.1 |
| $WT(btm-bound)$ | 0.3 | 0.6 | 0 | 0.6 | 0.6 | 0.3 | 0.1 | 0.1 | 0.1 | 0 | 0 | 0 | 0.1 | 0.3 |
| $WT(sht)$ | 1.5 | 2.1 | 1.5 | 1.6 | 2.1 | 0.8 | 0.6 | -0.3 | 0.4 | 0 | -0.1 | -0.1 | 0.3 | 0.3 |
| $WT(wnd)$ | 0.5 | 1.1 | 0.5 | 1.6 | 1.1 | 1.8 | 1.6 | 0.7 | 0.4 | 0 | 0.9 | 0.9 | 1.3 | 1.3 |
| $WT(h-sibl)$ | 0.6 | 0.6 | 0.6 | 0.6 | 0.6 | 0.6 | 0.6 | 0 | 0 | 0 | 0.6 | 0 | 0 | 0 |
| $WT'(sht)$ | 2.1 | 2.7 | 2.1 | 2.2 | 2.7 | 1.4 | 1.2 | -0.3 | 0.4 | 0 | 0.5 | -0.1 | 0.3 | 0.3 |
| $WT'(wnd)$ | 1.1 | 1.7 | 1.1 | 2.2 | 1.7 | 2.4 | 2.2 | 0.7 | 0.4 | 0 | 1.5 | 0.9 | 1.3 | 1.3 |

The result is reasonably good. All the real shutter areas get higher $WT'(sht)$ (namely, 2.1, 2.7, 2.1, 2.2 and 2.7) than non-shutter areas. Also, two lower window areas get high $WT'(wnd)$ values 2.4 and 2.2. However, a similar problem exists as in the Bayesian method, in that $WT'(wnd)$ for the shutter LMS is 2.2 which is as high as the

$WT'(wnd)$ for one of the windows LRW.

## 5. A Graphical Comparison

Figure 3 consists of histograms that depict the final belief on the hypothesis *window* derived with three different reasoning mechanisms. Data from Tables 1, 2 and 3 are used. As explained before, 14 areas are candidate window or shutter areas in the image 'House.sri'. For ease of viewing, the real shutters and windows are represented by boxes in different shades. The horizontal axes indicate the belief values, probabilities, or weights. Since no normalization is attempted, the magnitudes of the $Bel'$, $P'$, and $WT'$ are not simply compared. The histograms take the minimum and maximum values (i.e., 0.034 and 0.509 for belief values, 0.095 and 0.495 for probabilities, -0.3 and 2.7 for weights) and display them with an approximately equal width, so that the discrimination power of the different reasoning methods can be shown by the horizontal distances, e.g., how far windows are separated from non-windows. A good reasoning process should be able to obtain the highest $Bel'(wnd)$ (or $P'(wnd)$, $WT'(wnd)$) for the two true window areas. In Fig. 3 (a) the two true window areas LLW and LRW have the highest $Bel'(wnd)$ values, and thus are separated from the 'false' window areas. Figure 3 (b) and (c) show that one shutter area is not separated from the true window areas.

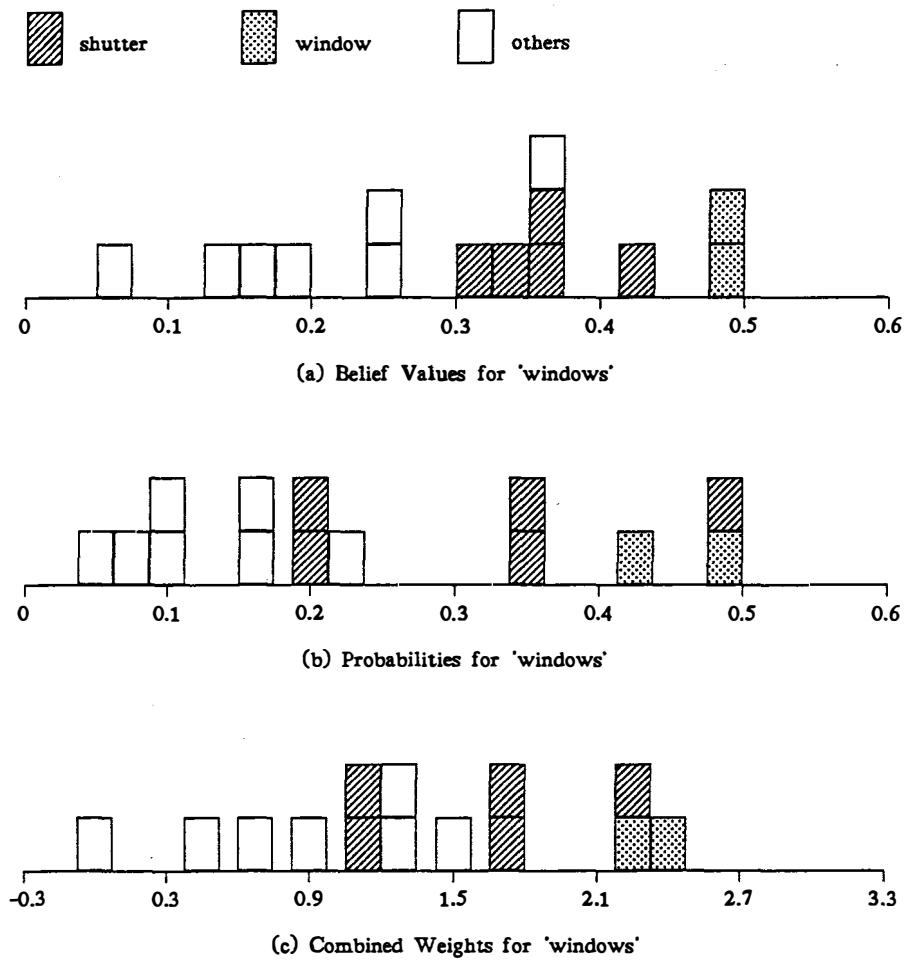

(a) Belief Values for 'windows'

(b) Probabilities for 'windows'

(c) Combined Weights for 'windows'

Fig. 3 A Graphical Comparison

Similar comparisons were also made in the analysis of another image 'Building1' [3]. The D-S mechanism was able to separate 12 real windows from 34 non-windows better than the Bayesian method and the simple weight combination method.

The weight combination method employed in this experiment uses simple addition and subtraction to combine weights. It treats all the features as of equal importance. In this aspect the Bayesian method has a behavior similar to that of the simple weight combination method. In contrast, our evidential reasoning approach incorporates the knowledge of *key features* and has a nice way to emphasize key features. For example, the mass distributions for the knowledge sources $m_{s_1}(sht)$ and $m_{s_1}(wnd)$ in section 2.2 all emphasize the role of the feature *elongation* by assigning more mass to it. This is an important factor that makes the evidential reasoning approach perform better in these examples.

From the examples described above, the comparison seems to be in favor of our evidential reasoning mechanism. At this point we are still studying these models and trying to find a way to deal with *key features* in the Bayesian implementation.

## 6. Conclusion

The belief functions can be used by an evidential reasoning mechanism for real-world image analysis. The Dempster Combination Rule offers a good tool for pooling multiple evidence. Our extended use of the Dempster-Shafer theory introduces a modular knowledge representation technique and its set-theoretic reasoning mechanism. Some preliminary experiments are illustrated in this paper. According to the comparison based on a limited number of images, it appears that our implementation and extension of D-S formalism makes effective use of *key features* and thus yields better result than the Bayesian method and the simple weight combining method.